\documentclass[conference]{IEEEtran}
\IEEEoverridecommandlockouts
\usepackage{cite}
\usepackage{amsmath,amssymb,amsfonts}
\usepackage{graphicx}
\usepackage{textcomp}
\usepackage{xcolor}
\usepackage[a4paper, total={184mm,239mm}]{geometry}
\def\BibTeX{{\rm B\kern-.05em{\sc i\kern-.025em b}\kern-.08em
    T\kern-.1667em\lower.7ex\hbox{E}\kern-.125emX}}

\usepackage{graphicx}
\usepackage{textcomp}
\usepackage{xcolor}
\usepackage{algorithm}
\usepackage{algpseudocode}
\usepackage{amssymb}
\usepackage{array}
\usepackage{hyperref}
\usepackage{booktabs}
\usepackage{caption}
\usepackage{color}
\usepackage{colortbl} 
\usepackage{comment}
\usepackage[inline]{enumitem}
\usepackage{epsfig}
\usepackage{etoolbox}
\usepackage{fancybox}
\usepackage{float}
\usepackage{fixltx2e}
\usepackage{graphicx}
\usepackage{hyperref}
\usepackage{listings}
\usepackage{multirow}
\usepackage{multicol}
\usepackage{placeins}
\usepackage{rotating}
\usepackage{setspace}
\usepackage[hang, tight]{subfigure}
\usepackage{threeparttable}
\usepackage{times}
\usepackage{url}
\usepackage{verbatim}
\usepackage{wrapfig}
\usepackage{xspace}
\usepackage{pifont}
\usepackage{mathrsfs} 
\usepackage{fancyhdr} 
\usepackage{orcidlink}

\begin{document}


\title{\textit{DynaMo}: Runtime Switchable Quantization for MoE with Cross-Dataset Adaptation}

\author{
\IEEEauthorblockN{Zihao Zheng}
\IEEEauthorblockA{\textit{School of Computer Science} \\
\textit{Peking University}\\
Beijing, China\\
zhengzihao@stu.pku.edu.cn\orcidlink{0009-0008-4624-2853}}
\and
\IEEEauthorblockN{Xiuping Cui}
\IEEEauthorblockA{\textit{School of Computer Science} \\
\textit{Peking University}\\
Beijing, China}
\and
\IEEEauthorblockN{Size Zheng}
\IEEEauthorblockA{\textit{School of Computer Science} \\
\textit{Peking University}\\
Beijing, China}
\and
\IEEEauthorblockN{Maoliang Li}
\IEEEauthorblockA{\textit{School of Computer Science} \\
\textit{Peking University}\\
Beijing, China}
\and
\IEEEauthorblockN{Jiayu Chen}
\IEEEauthorblockA{\textit{School of Computer Science} \\
\textit{Peking University}\\
Beijing, China}
\and
\IEEEauthorblockN{Yun Liang}
\IEEEauthorblockA{\textit{School of Integrated Circuits} \\
\textit{Peking University}\\
Beijing, China}
\and
\IEEEauthorblockN{Xiang Chen$^{\dagger}$}
\IEEEauthorblockA{\textit{School of Computer Science} \\
\textit{Peking University}\\
Beijing, China\\
xiang.chen@pku.edu.cn}
}
\maketitle

\begin{abstract}

As the Mix-of-Experts (MoE) architecture increases the number of parameters in large models, there is an even greater need for model quantization.
However, existing quantization methods overlook the expert dynamics of MoE across multiple datasets.
Moreover, the existing static quantization cannot adapt MoE to various data change scenarios.
In this paper, we perform a multi-level analysis to reveal MoE dynamics and define the significance of each channel/each expert.
Based on the analysis results, we propose \textit{DynaMo}, an end-to-end MoE quantization framework.
\textit {DynaMo} adopts an expert-level mixed-precision baseline quantization strategy, which ensures the quantized MoEs are compatible with multiple existing datasets. 
Furthermore, \textit {DynaMo} incorporates a channel-level dynamic switching mechanism to adapt these quantized MoE models to novel datasets.
Experiments show that \textit{DynaMo} achieves a 2.78$\sim$4.54 PPL decrease and a 1.85\%$\sim$3.77\% accuracy improvement in various datasets, with $\sim$3$\times$ inference speedup and negligible overhead.
\end{abstract}

\section{\textbf{Introduction}}
As Artificial Intelligence advances into the era of LLMs, their applications have expanded exponentially.
Since LLMs’ parameter density fails to adapt to the increasingly large and diverse datasets for their processing, the Mix-of-Experts (MoEs) architecture has become a promising solution~\cite {MoE-survey, switch-transformer}.
Each layer in MoE comprises multiple "expert" models. 
Trained to adapt to diverse data, these experts enable MoE to fit a wide range of data with strong performance.
In inference scenarios, MoE dynamically selects and sparsely activates a subset of these experts to fit the corresponding datasets.~\cite{moe-survey-data-perspective, HMOE, HoME}.

Despite their superior parameter scalability and memory efficiency via sparse expert activation, MoEs still require model compression methods~\cite{FedMoE, DA-MoE, MoE-efficient-inference}.
Among exsiting model compression methods, quantization is most effective: it reduces model size and accelerates computations by using low-precision parameter representations~\cite{Quant-survey}.

With advances in quantization techniques, methodology has shifted from parameter formats to the mapping between weights and complex input datasets. 
Methods like GPTQ~\cite{GPTQ} and QuantEase~\cite{QuantEase} use dataset analysis to establish data-weight mappings and compensate weights during quantization.
Subsequent approaches such as SmoothQuant~\cite{SmoothQuant} and AWQ~\cite{AWQ} further analyze outliers in specific datasets and transfer their scales to weights via data-weight mappings to minimize quantization loss.
Later methods including RPTQ~\cite{RPTQ} and Atom~\cite{Atom} leverage these mappings for finer-grained segmentation or reordering, then apply mixed-precision quantization to optimize the balance between compression rate and quantization loss.

\begin{figure*}[!t]
	\centering
    \includegraphics[width=7in]{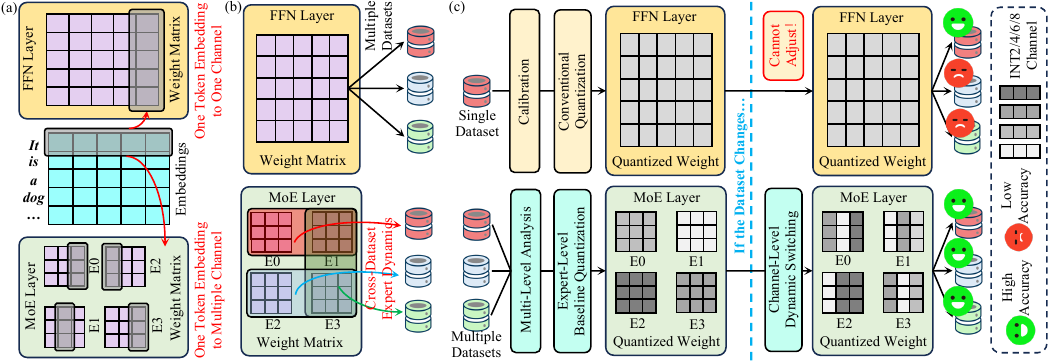}
    \vspace{-1mm}
	\caption{(a) MoE's One-to-Many Data-Weight Mappings (b) MoE's Expert/Weight Dynamics across Multiple Datasets (c) An Brief Overview of the Proposed \textit{DynaMo} Quantization Framework \vspace{-5mm}}
	\label{fig:1}
\end{figure*}


However, these strategies, designed for dense LLMs, yield suboptimal performance when applied to MoEs, for the following reasons:
(1) MoE architecture modifies data-weight mappings: as shown in Fig.~\ref{fig:1}~(a), each token embedding is dispatched to multiple channel weights across different experts, resulting in one-to-many correlations. By contrast, quantization methods designed for large language models (LLMs) only account for one-to-one correlations, failing to capture the complex data-weight mappings inherent to MoEs.
(2) MoE has inherent expert/weight dynamics: as shown in Fig.~\ref{fig:1}~(b), MoE adapts to a wide variety of datasets through a flexible combination of experts, resulting in scalable parameters and improved performance compared to LLM. Existing quantization methods are usually calibrated on one dataset, ignoring the dynamics of MoEs' expert/weight dynamics.

Recent studies (e.g., MoQE~\cite{MoQE}, MoEPTQ~\cite{MoE-Quantization}) have explored extending dense LLM quantization to MoEs, but they overlook the aforementioned characteristics of MoEs.
Moreover, the quantified MoE derived from these strategies lacks the adaptability to multi-data scenarios, a characteristic that contradicts the original design intention of MoE.
Therefore, it is crucial to conduct a thorough analysis of the complex mappings and intrinsic dynamics of MoEs, and to further rethink the quantization methods for MoE across multiple datasets.

In this paper, we conduct a multi-level analysis of MoEs' data-weight mappings and expert/weight dynamics, building metrics to capture the dynamics and expert/weight significance.
Based on the analysis results, we propose an end-to-end MoE quantization framework called \textit{DynaMo}, as shown in Fig.~\ref{fig:1}~(c).
Specifically, \textit{DynaMo} adopts an expert-level mixed-precision baseline quantization strategy, which ensures the quantized MoEs are compatible with multiple existing datasets.
Furthermore, \textit {DynaMo} incorporates a channel-level dynamic switching mechanism to adapt these quantized MoE models to novel datasets.
Overall, our contributions are as follows:
\begin{itemize}[leftmargin=*]
	\item We propose a multi-level analysis method for MoEs, quantitatively defining expert/weight significance within a single dataset and profiling its cross-dataset dynamics.
	\item Based on the analysis results, we propose \textit{DynaMo}, an end-to-end quantization framework for MoEs, which contains both expert-level and channel-level strategies, ensuring the performance across multiple datasets.
    \item We implement \textit {DynaMo} based on existing hardware and software, and conducted targeted optimizations. We further compare \textit {DynaMo} with SOTA quantization methods, analyzed its scalability and associated overhead in detail.
\end{itemize}

Experimental results show that, compared to SOTA methods (i.e., GPTQ~\cite{GPTQ} and MoEPTQ~\cite{MoE-Quantization}), \textit{DynaMo} achieves lower PPL (2.78$\sim$4.54) and higher accuracy (1.85\%$\sim$3.77\%) on various tasks.
Overall, the proposed \textit{DynaMo} delivers superior quantization performance with minimal overhead and suits diverse datasets and MoEs.
\section{\textbf{Preliminary}}
\label{tex: background_and_motivation}

\subsection{\textbf{MoE Model}}
\label{tex:back_1}
\textbf{Architecture and Mechanism.}
As illustrated in Fig.~\ref{fig:1}~(a), dense LLMs are structured as a series of interconnected blocks.
While in MoEs, the Feed-Forward Network (FFN) is replaced with multiple expert models~\cite{switch-transformer, JetMoE}.
Conventional expert selection depends on the Top-K mechanism, which selects a fixed number of experts per layer~\cite{MoE-survey, OLMoE}. 
Some studies also adopt soft or dynamic expert utilization strategies, activating different numbers of experts according to the characteristics of various data~\cite{keeptopk-1, keeptopk-2, AdapMoE}.

\textbf{Data-Weight Mappings.}
MoEs' architecture and mechanism modifies data-weight mappings.
Unlike LLMs, which use a single FFN to establish one-to-one mappings between channel weights and individual token embeddings, MoEs employ multiple experts. 
This design results in each token embedding corresponding to multiple channels across different experts.

\vspace{-1.5mm}
\subsection{\textbf{Model Quantization}}
\label{tex:back_2}
\textbf{Dense LLM Quantization.}
Since the rise of LLMs, quantization methods have shifted focus from parameter formats to dataset analysis to minimize accuracy loss.
GPTQ~\cite{GPTQ} and QuantEase~\cite{QuantEase} establish data-weight mappings to compensate weights during quantization, curbing accuracy drops. 
SmoothQuant~\cite{SmoothQuant} and AWQ~\cite{AWQ} target data outliers, leveraging such mappings for numerical smoothing or precision mixing to facilitate quantization. 
RPTQ~\cite{RPTQ} and Atom~\cite{Atom} further segment and reorder weight matrices based on data-weight correlations, developing channel-wise mixed-precision quantization for enhanced performance. 
These dataset-aware quantization methods achieves good performance.

\textbf{Challenge of MoE Quantization.}
In MoE, the expert selection process is inherently dataset-dependent, leading to complex data-weight mappings and dynamic expert/weight dynamics across different datasets.
This dynamic behavior presents a fundamental mismatch with static quantization schemes, which typically assume a fixed dataset. 
As a result, applying statically quantized MoE models to new datasets often leads to degraded accuracy due to MoE's dynamics.

\section{\textbf{Multi-Level MoE Dynamics Analysis}}
\label{tex:analysis}

During MoE inference, each token’s parsing activates multiple experts' channel weights, which makes it challenging to reveal the specific significance of each channel in each expert.
Moreover, MoE’s token-expert mappings cover both expert-level dispatching and intra-expert channel activations, making it hard to evaluate expert importance solely through token routing frequency or isolated activation statistics.
Thus, we propose a multi-level analysis to characterize MoE dynamics and better assess the significance of each channel and expert.

\vspace{-1.5mm}
\subsection{\textbf{Dataset-Specific Expert Weight Significance}}
\label{tex:analysis_1}
\begin{figure}[!t]
	\centering	
    \includegraphics[width=3.3in]{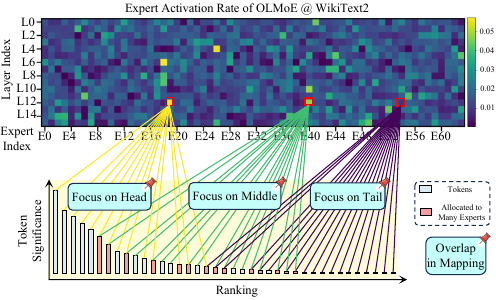}
    \vspace{-2mm}
	\caption{Data-Weight Mappings of MoEs \vspace{-6mm}}
	\label{fig:2}
\end{figure}

Since each token in the dataset is assigned to multiple experts in MoE, the token utilization of MoE differs from that of LLMs.
Thus, for a given dataset, we calculate the utilization rate of each token, and then establish a dataset-specific token ranking, as shown at the bottom of Fig.~\ref{fig:2}.
Based on this token ranking, we randomly select tokens from the provided dataset and perform MoE inference until the entire dataset is traversed.
During this process, we use a hook function to get the output logits of each channel weight in MoEs.
We perform a fine-grained significance analysis regarding each channel weight in each expert, namely, the mapping correlation to each token.
After that, we conducted a statistical analysis of the mapping to summarize the significance of each channel weight corresponding to the tokens in the given dataset.

\vspace{-1.5mm}
\subsection{\textbf{Cross-Dataset Expert Significance Dynamics}}
\label{tex:analysis_2}

By analyzing the significance of channel weights in each expert based on token utilization, we can determine the importance of each expert model under a single dataset. 
However, the importance of each expert depends on dataset-specific token utilization, and this utilization changes as the dataset varies. 
In such cases, traditional quantization methods, which rely on calibration with a single dataset, cannot ensure universality. 
Instead, these methods require tracking dataset changes to perform recalibration and quantization, a process that introduces substantial overhead. 
Therefore, we aim to develop a relatively general method for quantifying dynamically variable MoE models. 
To achieve this goal, we need to explore how experts behave dynamically as the dataset changes.

Based on the significance of channel weights, we propose "expert significance" to quantify and reflect the importance of each expert within a single dataset.
Specifically, assume each layer in MoE contains $N$ experts. 
The given dataset has $T$ tokens.
$\mathcal{W}_{\textnormal{ch}}$ means the channel weight.
And $\mathcal{S}_{\textnormal{ch}}(\cdot)$ are the significance of channel weights. 
Expert significance can be computed via Eq.~(\ref{eq:3-4}).
\vspace{-1mm}
\begin{equation}
\Bigg \{ \mathcal{S}^{j}_{\textnormal{exp}} = \frac{\sum_{i=1}^{t} \big \{ \mathcal{S_{\textnormal{ch}}} \big ( \sum_{k=1}^{K} \mathcal{W}_{\textnormal{ch}}^k \ | \ \textnormal{token} = \tau^{i} \big ) \big \} }{ \sum_{i=1}^{T} \big \{ \mathcal{S}_{\textnormal{ch}} \big ( \sum_{k=1}^{K} \mathcal{W}_{\textnormal{ch}}^k \ | \ \textnormal{token} = \tau^{i} \} \big )} \Bigg \}_{j=1}^{N},
\label{eq:3-4}
\end{equation}
\vspace{-3mm}

\begin{figure}[t]
	\centering
	\includegraphics[width=3.3in]{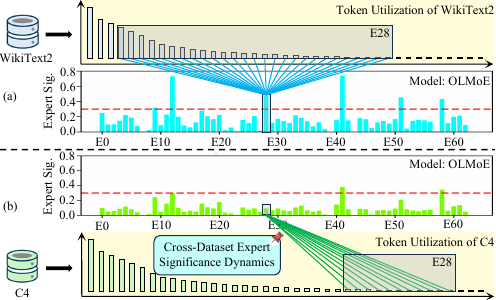}
    \vspace{-1mm}
	\caption{Cross-Dataset Expert Significance Dynamics \vspace{-5mm}}
	\label{fig:3}
\end{figure}

Subsequently, we calculate the expert significance of the same MoE layer across different datasets (WikiText2~\cite{WikiText2} and C4~\cite{C4}).
The results are shown in Fig.~\ref{fig:3}).
Under WikiText2 dataset, the 28\(^{\textnormal{th}}\) expert exhibits high significance ($>$0.5), while under C4 dataset, its significance is very low ($<$0.1).
The proposed expert significance metric effectively reveals the dynamics of experts in MoEs across different datasets.

\vspace{-2mm}
\subsection{\textbf{Synthesized Expert Baseline Significance \\\-\hspace{45mm} with Cross-Dataset Dynamics}}
The analysis presented above has elucidated the dynamics of expert significance across different datasets; however, these dynamics remain isolated from one another.
Specifically, these dynamics exhibit a strong correlation with individual datasets and cannot be unified across different data sources.
To adapt MoE to multiple datasets, developing an index capable of characterizing an expert’s synthesized performance across these diverse datasets is essential.
Therefore, we collect the expert significance under various datasets and fit them to a joint distribution, as Eq.~\ref{eq:3-5} shows.
$\mathcal{J}$ means the joint distribution. 
$\mathcal{D}$ means the dataset, and $\{ \mathcal{S}^{j}_{\textnormal{exp}} \}_{j=1}^{N}  \ \big \| \ {\mathcal{D}_{1}} $ means the expert significance under the first dataset.
\vspace{-1mm}
\begin{equation}
\mathcal{J} \stackrel{\textnormal{Fit}}{\longleftarrow} \Big ( \big \{ \mathcal{S}^{j}_{\textnormal{exp}} \big \}_{j=1}^{N} \ \big \| \ {\mathcal{D}_{1}},\cdots, \big \{ \mathcal{S}^{j}_{\textnormal{exp}} \big \}_{j=1}^{N} \ \big \| \ {\mathcal{D}_{n}} \Big ).
\label{eq:3-5}
\end{equation}
\vspace{-4mm}

\begin{figure}[b]
	\centering
    \vspace{-5mm}
	\includegraphics[width=3.3in]{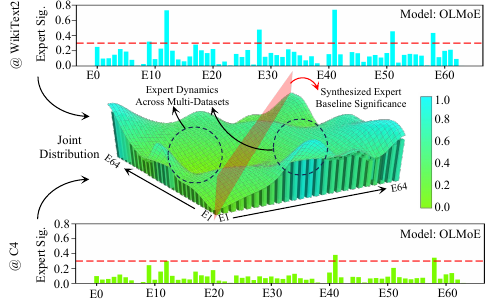}
	\caption{Joint Distribution of the Expert Significance}
	\label{fig:4}
    \vspace{-2mm}
\end{figure}

Fig.~\ref{fig:4} illustrates the results of the joint distribution. The diagonal bars in this joint distribution represent the synthesized expert significance across multiple datasets, whereas the remaining bars depict the dynamics of expert significance. 
This method yields a straightforward synthesized indicator for evaluating expert dynamics across multiple datasets, which can be directly applied to our quantization design.

\section{\textbf{Dataset-Aware MoE Quantization}}
\label{tex:design}

This section introduces a dataset-aware MoE quantization framework, called \textit{DynaMo}. 
It first performs expert-level mixed-precision static quantization (strategy determined by synthesized expert significance across datasets), then maps expert dynamics to channels and incorporates a fine-grained quantization switching mechanism.

\vspace{-0.5mm}
\subsection{\textbf{Expert-Level Baseline Quantization}}
\label{tex:design_1}
Due to the software and hardware limitations, existing quantization precision is mainly focused on four types (INT2, INT4, INT6, and INT8)~\cite{Quant-survey}. 
Although we can simply partition the synthesized baseline significance to four types, in fact, different baseline significance distributions still require us to accurately delineate the boundaries.

\textbf{Expert Baseline Significance Mapping with Quantization Segmentation.} Based on the synthesized expert baseline significance, we cluster the experts into four categories using the proposed Alg.~\ref{alg:1}.
We compute the number of tokens corresponding to each expert, and then initialize the clustering score ($U$) as the proportion of these expert-specific tokens relative to the total number of tokens in the dataset. 
Subsequently, we iteratively update the cluster centers by leveraging the product of dataset correlation and expert baseline significance.
We employ the cluster center $V$ and clustering score $U$ as evaluation metrics to determine the quantization precision for each cluster.
Specifically, the cluster with the highest $U$-value corresponds to the experts of greatest significance and is thus assigned the highest precision (INT8). Conversely, the cluster with the lowest $U$-value is assigned the lowest precision (INT2).

\begin{algorithm}[!b]
\caption{Clustering Algorithm based on Synthesized Expert Baseline Significance}
\footnotesize
\begin{algorithmic}[1]
\Require Synthesized expert baseline significance $\mathbf{X}=\{x_1, x_2, \ldots, x_n\}$, Number of clusters $c=4$, Fuzzifier $m>1$, Tolerance $\epsilon$
\Ensure Cluster centers $\mathbf{V} = \{v_1, v_2, \ldots, v_c\}$, Cluster scores $\mathbf{U}$
\State Initialize $\mathbf{U}$ with $u_{ik} \in [0,1]$ and $\sum_{k=1}^{c} u_{ik} = 1$ for all $i$
\Repeat
    \For{$k = 1, \ldots, c$} $v_k \Leftarrow {\sum_{i=1}^{n} u_{ik}^m x_i} / {\sum_{i=1}^{n} u_{ik}^m}$ \EndFor
    \For{$\forall i,k$} $u_{ik} \Leftarrow 1 / {\sum_{j=1}^{c} \left( \frac{\|x_i - v_k\|}{\|x_i - v_j\|} \right)^{\frac{2}{m-1}}}$ \EndFor
    \Until $\|\mathbf{U}^{new} - \mathbf{U}^{old}\| < \epsilon$
    \Return $\mathbf{V}$ and $\mathbf{U}$
\end{algorithmic}
\label{alg:1}
\end{algorithm}

\begin{figure}[!t]
	\centering
	\includegraphics[width=3.3in]{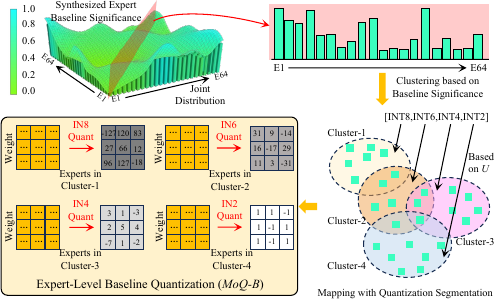}
    \caption{Expert-level Mix-Precision Baseline Quantization\vspace{-5mm}}
	\label{fig:5}
\end{figure}

\textbf{Quantization for Expert Located in Clustering Boundary.}
As illustrated in Fig.~\ref{fig:5}, the clustering results exhibit overlap across different categories. 
For experts situated in these overlapping regions, we adopt a low-precision-first strategy.
The strategy is supported by experimental results, which demonstrate that the precision of these experts exerts a negligible impact on the loss of quantization accuracy. 0
Applying low-bit quantization to these experts effectively increases the overall compression ratio.
Therefore, experts in the overlapping regions are assigned to the low-precision cluster for quantization. 

After determining the quantization precision for each cluster, we perform mix-precision quantization at the expert level.
Given computational complexity considerations, we adopt uniform quantization to reduce the computational overhead of the dequantization process.
It is important to note that all aforementioned analyses, along with the baseline quantization process, are conducted offline and do not compromise the inference speed of the quantized MoE model.

\vspace{-1.5mm}
\subsection{\textbf{Channel-Level Dynamic Quantization Switching}}
\label{tex:design_2}

\begin{figure}[!t]
	\centering
	\includegraphics[width=3.3in]{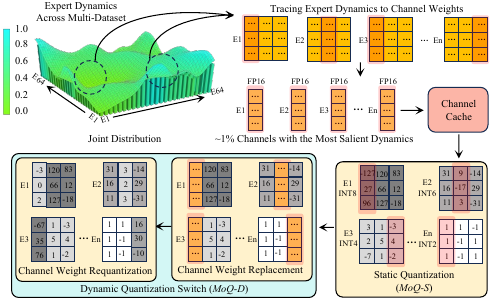}
	\caption{Channel-level Dynamic Quantization Switching\vspace{-5mm}}
	\label{fig:6}
\end{figure}

In the preceding section, we discussed baseline precision determination and the adjustment of key boundary experts during quantization across multiple datasets.
However, applying dynamic switching to all expert parameters would lead to prohibitive optimization costs. 
As analyzed in Section~\ref{tex:analysis}, MoE dynamics originate from the mapping between channel weights and tokens. 
Tracing these dynamics to individual channel weights, we find only $\sim$1\% of channels in each expert exhibit the most prominent dynamics.
This observation is also supported by existing research~\cite{AWQ}.
Dynamic channel switching in response to dataset changes allows the model to attain near-optimal quantization performance with minimal computational overhead. 
Yet, this requires addressing two key aspects: channel replacement and precision redetermination.

\textbf{Channel Weights Replacement based on Caching.}
The model undergoing baseline quantization suffers from irrecoverable quantization loss across each channel. 
Consequently, we are unable to restore the quantized channel weights to a higher precision; we can only adjust them to a lower precision.
To enable flexible adjustment of channel weight precision, the original unquantized FP16 channels must be retained.
As shown in the upper right part of Fig.~\ref{fig:6}, we cache $\sim$1\% of the channels with the most salient dynamics.
At the same time, to facilitate rapid searching, we also cache their channel indexes and the indexes of the corresponding experts.
As shown in the lower right part of Fig.~\ref{fig:6}, these cached indexes allow fast and accurate retrieval of target channels in the quantized MoE.
We then replace the quantized channels with the cached FP16 channels whenever the indexes match.
The replaced FP16 channels provide the basis for quantization switching, overcoming the limitation of upward precision adjustment.

\textbf{Channel Weights Requantization for New Datasets.}
After channel weight replacement, we calculate the expert significance for the new dataset ($\{ \mathcal{S}^{j}_{\textnormal{exp}} \}_{j=1}^{N}  \ \big \| \ {\mathcal{D}_{\textnormal{new}}} $) based on Eq.~\ref{eq:3-4}.
Then, we apply Alg.~\ref{alg:1} to cluster the expert significance for the new dataset ($\{ \mathcal{S}^{j}_{\textnormal{exp}} \}_{j=1}^{N}  \ \big \| \ {\mathcal{D}_{\textnormal{new}}}$) and compare this result with the clustering results in Fig.~\ref{fig:5}.
Based on the differences between the two clustering results, we can determine the specific precision to which the FP16 channels within each expert should be requantized.
For instance, under the joint distribution, the 28\(^\textnormal{th}\) expert belongs to Cluster-1 (INT8); however, with the new dataset, this 28\(^\textnormal{th}\) expert is reassigned into Cluster-3 (INT4). 
In this case, the FP16 channels of the 28\(^\textnormal{th}\) expert should be requantized to the INT4 format.
Following this approach, we requantize all FP16 channels to the appropriate precision.
It is worth noting that the precision of these channels after requantization may be either higher or lower than before, thereby allowing fine-tuning of each expert’s performance, with a negligible overhead.
\section{\textbf{Quantization Framework Implementation}}
\label{tex:implementation}

\begin{figure}[!t]
	\centering
	\includegraphics[width=3.3in]{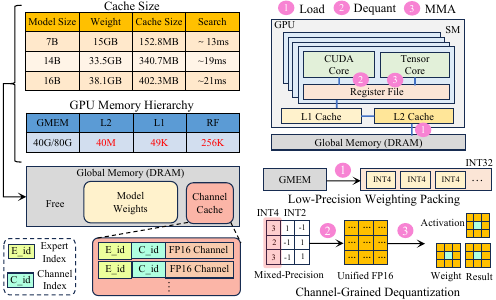}
	\caption{Implementation Details of \textit{DynaMo} \vspace{-5mm}}
	\label{fig:7}
\end{figure}

Due to the expert-level mixed precision and channel-level dynamic switching mechanism
, \textit{DynaMo} encounters challenges in hardware compatibility and computational efficiency.
This section will detail how \textit{DynaMo} addresses these challenges.



\vspace{-1.5mm}
\subsection{\textbf{Quantization Switch based on Channel Weight Caching}}
\textbf{Storage Consideration for the Channel Cache.}
Fig.~\ref{fig:7} presents the cache size and search time required for common MoE models (7B$\sim$16B).
Among GPU memory components, only global memory (GMEM) accommodates cache, so we allocate a dedicated GMEM space for the channel cache.
As we select merely 1\% of channels, the cache size is negligible. 
For example, in a 7B MoE, the cache requires only 152.8MB, equivalent to 1.02\% of the MoE’s weights storage.
Notably, the processes of constructing the channel cache and calculating expert significance come before quantization adjustment, ensuring no additional overhead.

\textbf{Data Format in the Channel Cache.}
We use the expert indexes and channel indexes as data headers and store them in the channel cache.
This method ensures that, during replacement, the corresponding FP16 Cache can be quickly retrieved based on the expert indexes and channel indexes, thereby significantly reducing the overhead of online dynamic adjustment.
\vspace{-2mm}
\subsection{\textbf{Kernel Optimization under Mix-Precision Quantization}}
\textbf{Quantized Weight Packing Load.}
We use packing to enable efficient loading of low-precision weights from GMEM to the GPU. 
Specifically, we pack multiple low-precision data formats into an INT32, allowing simultaneous loading of multiple weights via original hardware instructions. 
Post-loading, we unpack the data within the GPU. 
This method indirectly improves hardware bandwidth utilization and constitutes the primary performance driver for inference acceleration.

\textbf{Channel-Grained Dequantization.}
Mixed-precision quantization may lead to computational challenges, as data in different formats cannot be directly processed.
To address this, we uniformly dequantize the loaded low-precision weights into FP16 using CUDA cores.
Both weights and activations in FP16 can then execute MMA on tensor cores, resolving the computational issue.
Furthermore, we fuse the dequantization and MMA kernels to enhance overall computational efficiency.
\section{\textbf{Experiments}}
\label{tex:experiments}

\begin{table}[!b]
    \centering
    \vspace{-3mm}
    \caption{Evaluations on Language Modeling Tasks.}
    \label{tab:6-1}
    \scriptsize
    \begin{tabular}{ccccccc}
    \toprule
    \toprule
    
    \textbf{Model} & \textbf{Method} & \textbf{\#Bits} & \textbf{Wiki} & \textbf{C4} & \textbf{Avg. ($\downarrow$)} & ~\\
    \midrule
    
    \multirow{3}*{OLMoE} & w/o Quant & 16 & 7.41 & 11.42 & 9.42 \\
    \cmidrule{2-6}
    ~ & GPTQ & 3 & 11.65 & 18.86 & 15.26 & ~ \\
    \multirow{2}*{1B / 7B} & MoEPTQ & 3.26 & 15.44 & 26.04 & 20.74 & ~ \\
    ~ & \cellcolor{gray!20} \textit{DynaMo} & \cellcolor{gray!20}2.95 & \cellcolor{gray!20}9.64 & \cellcolor{gray!20}15.31 & \cellcolor{gray!20}12.48 & \textcolor{blue}{$\downarrow$ 2.78} \\
    \midrule
    
    \multirow{3}*{MoE-Girl} &
        w/o Quant & 16 & 8.43 & 13.13 & 10.78 & ~ \\
    \cmidrule{2-6}
    ~ & GPTQ & 3 & 12.77 & 21.38 & 17.08 & ~ \\
    \multirow{2}*{1B / 7B} & MoEPTQ & 3.26 & 16.88 & 29.40 & 23.14 & ~ \\
    ~ & \cellcolor{gray!20}\textit{DynaMo} & \cellcolor{gray!20}2.89 & \cellcolor{gray!20}10.47 & \cellcolor{gray!20}17.87 & \cellcolor{gray!20}14.17 & \textcolor{blue}{$\downarrow$ 2.91} \\
    \midrule
    
    \multirow{3}*{Qwen1.5} & w/o Quant & 16 & 7.02 & 10.03 & 8.53 & ~ \\
    \cmidrule{2-6}
    ~ & GPTQ & 3 & 10.99 & 20.84 & 15.92 & ~ \\
    \multirow{2}*{3B / 14B} & MoEPTQ & 3.35 & 9.93 & 18.49 & 14.21 & ~\\
    ~ & \cellcolor{gray!20}\textit{DynaMo} & \cellcolor{gray!20}3.05 & \cellcolor{gray!20}8.51 & \cellcolor{gray!20}14.24 & \cellcolor{gray!20}11.38 & \textcolor{blue}{$\downarrow$ 4.54} \\
    \midrule
    
    \multirow{3}*{DS-MoE} & w/o Quant & 16 & 7.36 & 9.22 & 8.29 & ~ \\
    \cmidrule{2-6}    
    ~ & GPTQ & 3 & 10.47 & 15.19 & 12.83 & ~ \\
    \multirow{2}*{3B / 16B} & MoEPTQ & 3.31 & 8.49 & 15.61 & 12.05 & ~ \\
    ~ & \cellcolor{gray!20}\textit{DynaMo} & \cellcolor{gray!20}2.98 & \cellcolor{gray!20}7.94 & \cellcolor{gray!20}11.49 & \cellcolor{gray!20}9.72 & \textcolor{blue}{$\downarrow$ 3.11} \\
    \bottomrule
    \bottomrule
    \end{tabular}
\end{table}

\begin{table*}[!t]
    \centering
    \caption{Evaluations on Zero-Shot Inference Tasks}
    \vspace{-1mm}
    \label{tab:6-2}
    \scriptsize
    \begin{tabular}{ccccccccccc}
    \toprule
    \toprule
    {\textbf{Model}} & \textbf{Method} & {\textbf{\#Bits}} & \textbf{ARC-challenge} & \textbf{ARC-easy} & \textbf{RTE} & \textbf{PIQA} & \textbf{COPA} & \textbf{CB} & \textbf{Avg. ($\uparrow$)} & ~\\
    \midrule
    
    \multirow{3}{*}{OLMoE} & w/o Quant & 16 & 29.69\% & 48.48\% & 54.51\% & 61.86\% & 71.00\% & 41.07\% & 51.10\% & ~\\
    \cmidrule{2-10}
    
    ~ & GPTQ & 3 & 25.34\% & 41.12\% & 51.99\% & 58.81\% & 62.00\% & 39.29\% & 46.43\% \\
    \multirow{2}{*}{1B / 7B} & MoEPTQ & 3.26 & 24.83\% & 38.38\% & 50.54\% & 56.86\% & 65.00\% & 42.86\% & 42.27\% & ~ \\
    
    
    ~ & \cellcolor{gray!20}\textit{DynaMo} & \cellcolor{gray!20}2.97 & \cellcolor{gray!20}25.85\% & \cellcolor{gray!20}43.52\% & \cellcolor{gray!20}54.15\% & \cellcolor{gray!20}58.87\% & \cellcolor{gray!20}65.00\% & \cellcolor{gray!20}46.43\% & \cellcolor{gray!20}48.94\% & \textcolor{blue}{$\uparrow$ 2.54\%} \\
    
    \midrule
    
    \multirow{3}{*}{MoE-Girl} & w/o Quant & 16 & 31.31\% & 50.84\% & 55.95\% & 62.62\% & 66.00\% & 41.07\% & 51.30\% \\
    \cmidrule{2-10}
    
    ~ & GPTQ & 3 & 25.17\% & 38.80\% & 55.59\% & 60.33\% & 62.00\% & 39.28\% & 46.86\% \\
    \multirow{2}*{1B / 7B} & MoEPTQ & 3.26 & 24.23\% & 37.04\% & 53.06\% & 57.88\% & 59.00\% & 41.07\% & 45.38\% \\
    
    
    ~ & \cellcolor{gray!20}\textit{DynaMo} & \cellcolor{gray!20}2.88 & \cellcolor{gray!20}26.02\% & \cellcolor{gray!20}44.87\% & \cellcolor{gray!20}53.43\% & \cellcolor{gray!20}61.32\% & \cellcolor{gray!20}62.00\% & \cellcolor{gray!20}44.64\% & \cellcolor{gray!20}48.71\% & \textcolor{blue}{$\uparrow$ 1.85\%} \\
    
    \midrule
    
    \multirow{3}*{Qwen1.5-MoE} & w/o Quant & 16 & 33.11\% & 51.30\% & 71.84\% & 72.47\% & 81.00\% & 25.01\% & 55.79\% & ~\\
    
    \cmidrule{2-10}
    
    ~ & GPTQ & 3 & 26.54\% & 39.44\% & 54.51\% & 63.87\% & 72.00\% & 24.76\% & 46.85\% \\
    \multirow{2}*{3B / 14B} & MoEPTQ & 3.36 & 24.23\% & 37.04\% & 53.07\% & 63.22\% & 67.00\% & 24.37\% & 44.82\% & ~\\
    
    
    ~ & \cellcolor{gray!20}\textit{DynaMo} & \cellcolor{gray!20}3.04 & \cellcolor{gray!20}27.13\% & \cellcolor{gray!20}43.69\% & \cellcolor{gray!20}55.23\%& \cellcolor{gray!20}68.72\% & \cellcolor{gray!20}75.00\% & \cellcolor{gray!20}33.93\% & \cellcolor{gray!20}50.62\% & \textcolor{blue}{$\uparrow$ 3.77\%} \\
    
    \midrule
    
    \multirow{3}*{DeepSeek-MoE} & w/o Quant & 16 & 40.61\% & 71.55\% & 54.51\% & 76.22\% & 82.00\% & 41.07\% & 60.99\% & ~ \\
    \cmidrule{2-10}
    
    ~ & GPTQ & 3 & 33.62\% & 62.04\% & 52.34\% & 73.94\% & 78.00\% & 44.64\% & 57.43\%\\
    \multirow{2}*{3B / 16B} & MoEPTQ & 3.36 & 31.99\% & 60.06\% & 53.43\% & 69.42\% & 77.00\% & 41.07\% & 55.50\% & ~ \\
    
    
    ~ & \cellcolor{gray!20}\textit{DynaMo} & \cellcolor{gray!20}2.99 & \cellcolor{gray!20}34.71\% & \cellcolor{gray!20}64.19\% & \cellcolor{gray!20}53.92\% & \cellcolor{gray!20}75.34\% & \cellcolor{gray!20}81.00\% & \cellcolor{gray!20}47.87\% & \cellcolor{gray!20}59.51\% & \textcolor{blue}{$\uparrow$ 2.08\%} \\
    \bottomrule
    \bottomrule
    \end{tabular}
    \vspace{-4mm}
\end{table*}

\vspace{-1.5mm}
\subsection{\textbf{Experiments Setup}}
To evaluate \textit{DynaMo}, we select GPTQ~\cite{GPTQ} and MoEPTQ~\cite{MoE-Quantization} as baselines.
We test \textit{DynaMo} on various MoEs~\cite{OLMoE, MoE-Girl, Qwen1.5-MoE, deepseek-moe}.
We test \textit{DynaMo} under various tasks and input data distributions, including WikiText2~\cite{WikiText2}, C4~\cite{C4}, ARC~\cite{ARC}, RTE~\cite{RTE}, PIQA~\cite{PIQA}, COPA~\cite{COPA}, and CB~\cite{CB}. 
All experiments are performed on NVIDIA A100 GPUs.


\vspace{-2mm}
\subsection{\textbf{Evaluations on Language Modeling \& Zero-Shot Tasks}}
We evaluate \textit{DynaMo} on multiple language modeling tasks.
Results in Tab.~\ref{tab:6-1} shows that at an average $\sim$3-bit precision, \textit{DynaMo} has a 2.78$\sim$4.54 decrease in perplexity (PPL) compared to baseline.
Next, we test \textit{DynaMo} on multiple zero-shot inference tasks.
As shown in Tab.~\ref{tab:6-2}, \textit{DynaMo} improves accuracy by 1.85\%$\sim$3.77\% over baselines.
These tests demonstrate that \textit {DynaMo} can adapt to various datasets while achieving lower quantization loss at the same compression ratio.

\begin{figure}[!t]
	\centering
    \vspace{-2mm}
	\includegraphics[width=3.3in]{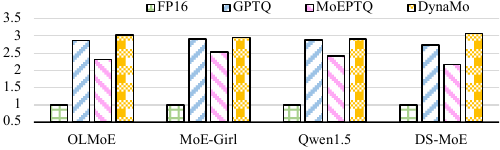}
	\caption{Speedup of \textit{DynaMo} on Various MoEs \vspace{-4mm}}
	\label{fig:8}
\end{figure}

\vspace{-1.5mm}
\subsection{\textbf{Results of DynaMo's Inference Speed}}
The inference speed of \textit{DynaMo} are shown in Fig.~\ref{fig:8}. 
Compared to FP16 model, \textit{DynaMo} achieve 2.91$\sim$3.08$\times$ speedup.
Compared with baselines, \textit{DynaMo} still shows slight speed improvements.
Existing research shows quantization’s primary performance boost stems from loading low-precision data from GMEM to the GPU. 
Though mixed precision adds extra computational overhead, we maximize bandwidth utilization by packing low-precision data during loading, enabling inference speed on par with SOTA methods.
Note that the key advantage of \textit {DynaMo} lies in its flexibility and adjustability when addressing variations across datasets. 
Its core focus is on guaranteeing accuracy across diverse datasets, rather than chasing speed.

\begin{table}[!b]
    \centering
    \vspace{-1.5mm}
    \caption{Ablation Study}
    \label{tab:6-3}
    \scriptsize
    \begin{tabular}{ccccc}
    \toprule
    \toprule
    \textbf{Model} & \textbf{Task} & \textbf{GPTQ} & \textbf{Only BQ} & \textbf{+ DQS} \\
    \midrule
    \multirow{2}{*}{OLMoE} & Wiki & 11.65 & 10.78~\textcolor{blue}{$\downarrow$ 0.87} & 9.64~\textcolor{blue}{$\downarrow\downarrow$ 1.23} \\
    ~ & CB & 39.29\% & 42.79\%~\textcolor{blue}{$\uparrow$ 3.50\%} & 46.43\%~\textcolor{blue}{$\uparrow\uparrow$ 3.64\%} \\
    \midrule
    
    \multirow{2}{*}{MoE-Girl} & Wiki & 12.77 & 11.23~\textcolor{blue}{$\downarrow$ 1.54} & 10.47~\textcolor{blue}{$\downarrow\downarrow$ 0.76}\\
    ~ & CB & 39.28\% & 42.66\%~\textcolor{blue}{$\uparrow$ 3.38\%} & 44.64\%~\textcolor{blue}{$\uparrow\uparrow$ 1.98\%}\\
    \midrule
    
    \multirow{2}{*}{Qwen1.5} & Wiki & 10.99 & 9.82~\textcolor{blue}{$\downarrow$ 1.17} & 8.51~\textcolor{blue}{$\downarrow\downarrow$ 1.31} \\
    ~ & CB & 24.76\% & 29.69\%~\textcolor{blue}{$\uparrow$ 4.93\%} & 33.93\%~\textcolor{blue}{$\uparrow\uparrow$ 4.24\%}\\
    \midrule
    
    \multirow{2}{*}{DS-MoE} & Wiki & 10.47 & 8.52~\textcolor{blue}{$\downarrow$ 1.95} & 7.94~\textcolor{blue}{$\downarrow\downarrow$ 0.58} \\
    ~ & CB & 44.64\% & 45.91\%~\textcolor{blue}{$\uparrow$ 1.27\%} & 47.87\%~\textcolor{blue}{$\uparrow\uparrow$ 1.96\%} \\
    \bottomrule
    \bottomrule
    \end{tabular}
\end{table}

\vspace{-1.5mm}
\subsection{\textbf{Ablation Study}}
We conducted an ablation study on \textit{DynaMo}, using WikiText2 for language modeling and CB for zero-shot tasks. 
First, we performed only expert-level baseline quantization (BQ) on the C4 and RTE datasets, then compared \textit{DynaMo} and GPTQ. 
Next, we added channel-level dynamic switching (DQS) based on WikiText2 and CB, and retested.
All results are shown in Tab.~\ref{tab:6-3}.
With Only BQ, PPL decreased by 0.87$\sim$1.95 and accuracy improved by 1.27\%$\sim$4.93\%, confirming our multi-stage analysis accurately captures MoE’s inherent dynamics, identifies optimal expert quantization precision, and mitigates quantization-induced accuracy loss. 
Adding +DQS further reduced PPL by 0.58$\sim$1.31 and boosted accuracy by 1.96\%$\sim$4.24\%, showing dynamic switching helps quantized MoE adapt to new datasets for better performance.

\begin{figure}[!t]
	\centering
    \vspace{-3.5mm}
	\includegraphics[width=3.3in]{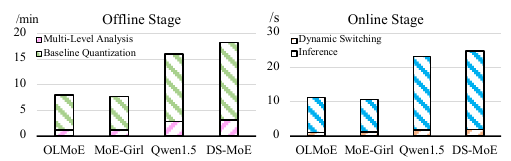}
	\caption{Offline/Online Stage Overhead of \textit{DynaMo}  \vspace{-5mm}}
	\label{fig:9}
\end{figure}
\vspace{-1.5mm}
\subsection{\textbf{Overhead Discussion}}
We evaluate \textit{DynaMo}'s overhead. 
Given that multi-level analysis and baseline quantization are offline processes, while dynamic quantization switching and model inference are online, we test them separately.
As shown in Fig.~\ref{fig:9}, multi-stage analysis accounts for 14.90\%$\sim$17.33\% of offline phase overhead, and dynamic quantization switching only takes up 7.73\%$\sim$10.68\% of online overhead. 
Note that inference time here refers to the duration for MoE to generate 1024 tokens; additionally, dynamic quantization switching is only needed when the dataset changes. Overall, \textit{DynaMo}'s cost is negligible.
\section{\textbf{Conclusion}}
\label{tex:conclusion}

This paper proposes \textit{DynaMo}, a novel MoE quantization framework.
First, it enables expert-level mixed-precision baseline quantization compatible with multiple existing datasets; then, it introduces channel-level dynamic switching to adapt quantized MoEs to new datasets.
Experiments show that \textit{DynaMo} outperforms SOTA works in quantization performance.

\bibliographystyle{IEEEtran}
\bibliography{ref/reference.bib}

\end{document}